\documentclass{article}

\usepackage{fullpage}
\usepackage{listings} 
\usepackage{graphicx} 
\usepackage{amsmath}
\usepackage{amssymb}
\usepackage{multirow}
\usepackage{hyperref}
\hypersetup{
    colorlinks=true,
    linkcolor=blue,
    filecolor=magenta,      
    urlcolor=blue,
    citecolor=black,
    pdftitle={A Translation of Probabilistic Event Calculus into Markov Decision Processes},
    pdfpagemode=FullScreen,
}
\usepackage[dvipsnames]{xcolor}

\definecolor{codegreen}{rgb}{0,0.6,0}
\definecolor{codegray}{rgb}{0.5,0.5,0.5}
\definecolor{codepurple}{rgb}{0.2,0,0.82}
\definecolor{backcolour}{rgb}{0.9,0.9,0.9}

\lstdefinestyle{mystyle}{
    backgroundcolor=\color{backcolour},   
    commentstyle=\color{codegreen},
    keywordstyle=\color{codepurple},
    numberstyle=\tiny\color{codegray},
    stringstyle=\color{codepurple},
    basicstyle=\ttfamily\footnotesize,
    breakatwhitespace=false,         
    breaklines=true,                 
    captionpos=b,                    
    keepspaces=true,                 
    numbers=left,                    
    numbersep=5pt,                  
    showspaces=false,                
    showstringspaces=false,
    showtabs=false,                  
    tabsize=2
}

\newcommand{\cardinalityof}[1]{\left\lvert #1 \right\rvert}

\newcommand{\powersetof}[1]{\mathcal{P}(#1)}
\newcommand{\indicatorfunc}[1]{\mathbf{1}\left( #1\right)}
\newcommand{\instants}{\mathcal{I}}
\newcommand{\mdptimes}{\mathcal{T}}

\newcommand{\vals}{\ensuremath{\mbox{\textit{vals}}}}
\newcommand{\pecdomain}{\mathcal{D}}
\newcommand{\numfluents}{n_F}
\newcommand{\numvalues}{n_V}
\newcommand{\numstates}{n_S}
\newcommand{\numinstants}{n_I}
\newcommand{\numvecreps}{n_X}
\newcommand{\numactions}{n_A}
\newcommand{\orderedelement}[2]{#1^{(#2)}}
\newcommand{\timemap}{\delta}
\newcommand{\pecfluent}{F}
\newcommand{\pecvalues}{\mathcal{V}}
\newcommand{\pecfluents}{\mathcal{F}}
\newcommand{\pecfluentstate}{\tilde{S}}
\newcommand{\pecfluentstates}{\mathcal{\tilde{S}}}
\newcommand{\pecstate}{S}
\newcommand{\pecstates}{\mathcal{S}}
\newcommand{\partialfluentstate}{\tilde{X}}
\newcommand{\partialfluentstates}{\mathcal{\tilde{X}}}
\newcommand{\valueindexof}{\iota_V}
\newcommand{\vectorrepset}{\mathcal{X}}
\newcommand{\vectorrepindexof}{\iota_X}

\newcommand{\stateindexof}{\iota_S}

\newcommand{\mdpstates}{\mathcal{S}}
\newcommand{\mdpactions}{\mathcal{A}}
\newcommand{\mdpinitdist}{p_{0}}
\newcommand{\transfunc}{T}
\newcommand{\pecaction}{U}
\newcommand{\pecactions}{\mathcal{U}}
\newcommand{\possibleactionsat}[1]{\pecactions_{#1}}
\newcommand{\actionsituations}{\mathcal{A}_u}
\newcommand{\actionsituation}{A_u}
\newcommand{\nullactionsituation}{A_\emptyset}
\newcommand{\nullaction}{a_\emptyset}
\newcommand{\situationindexof}{\iota_U}
\newcommand{\numsituations}{n_U}
\newcommand{\encodedsetsituations}{\mathcal{\hat{A}}}

\newcommand{\bodyactions}{A_\theta}
\newcommand{\bodystates}{\tilde{\mathcal{S}}_\theta}

\newcommand{\query}{Q}
\newcommand{\cond}{C}

\newcommand{\literal}{L}
\newcommand{\literals}{\mathcal{L}}

\title{A Translation of Probabilistic Event Calculus into Markov Decision Processes}

\author{
  Lyris Xu\thanks{Department of Information Studies, University College London, UK. \texttt{l.xu24@ucl.ac.uk}}
  \and
  Fabio Aurelio D'Asaro\thanks{Dipartimento di Studi Umanistici, Università del Salento, Italy. \texttt{fabioaurelio.dasaro@unisalento.it}}
  \and
  Luke Dickens\thanks{Department of Information Studies, University College London, UK. \texttt{l.dickens@ucl.ac.uk}}
}

\begin{document}

\maketitle

\begin{abstract}
Probabilistic Event Calculus (PEC) is a logical framework for reasoning about actions and their effects in uncertain environments, which enables the representation of probabilistic narratives and computation of temporal projections. The PEC formalism offers significant advantages in interpretability and expressiveness for narrative reasoning. However, it lacks mechanisms for goal-directed reasoning. This paper bridges this gap by developing a formal translation of PEC domains into Markov Decision Processes (MDPs), introducing the concept of ``action-taking situations'' to preserve PEC's flexible action semantics. The resulting PEC-MDP formalism enables the extensive collection of algorithms and theoretical tools developed for MDPs to be applied to PEC's interpretable narrative domains. We demonstrate how the translation supports both temporal reasoning tasks and objective-driven planning, with methods for mapping learned policies back into human-readable PEC representations, maintaining interpretability while extending PEC's capabilities.
\end{abstract}

\section{Introduction}

Reasoning about actions and their effects in dynamic, uncertain environments is a fundamental challenge in Artificial Intelligence (AI). The importance of narrative reasoning-- the ability to represent and reason about sequences of events and their causal relationships over time-- has been recognised since the early days of AI, leading to various temporal reasoning formalisms \cite{McCarthy1969-MCCSPP, allen1983maintaining, kowalski1986logic, sandewall1992features}. 
Building on these early foundations, the Probabilistic Event Calculus (PEC) \cite{d2017foundations} has emerged as a powerful framework for representing and reasoning about uncertain scenarios. PEC extends the Event Calculus (EC) \cite{kowalski1986logic, miller2002some}, to provide an ``action-language'' style framework for modelling actions, their effects, and the evolution of world states over time under uncertainty. 

PEC-style frameworks offer highly interpretable and flexible representations of complex narratives, with demonstrated applications in domains such as medicine, environmental monitoring, and commonsense reasoning \cite{d2020probabilistic, d2017foundations}. These frameworks support temporal reasoning tasks like temporal projection-- computing the probability of future state properties given past actions and states. Several implementation approaches exist, including Answer Set Programming \cite{d2017foundations}, approximate sampling with Anglican \cite{dasaro2019phd}, and iterative temporal modelling \cite{d2023application}. 

Markov Decision Processes (MDPs), meanwhile, have established themselves as a powerful framework for modelling time-evolving systems controlled by an agent. MDPs and their variants are widely used across AI for decision-making under uncertainty and serve as the foundation for many statistical and reinforcement learning algorithms. While MDPs excel at control optimisation problems, PEC and its variants offer superior narrative interpretability but lack mechanisms for learning goal-directed behaviour.

A translation between these frameworks presents a compelling opportunity to bridge this gap, combining PEC's human-readable representation with MDP's computational efficiency and reinforcement learning capabilities. Such a bridge would allow for both efficient PEC implementation and the application of statistical learning techniques to narrative reasoning tasks. Towards this goal, we have developed a novel translation of PEC into an MDP framework, termed the PEC-MDP. This translation preserves the core assumptions and semantics of PEC while enabling the application of a wide range of MDP-based algorithms to PEC's human-readable domains.

The key contributions of this work include:
\begin{itemize}
    \item A formal translation of PEC domains to PEC-MDPs, with a Python implementation made available through a shared repository.
    \item An approach for performing temporal projection via the PEC-MDP formalism.
    \item A novel approach to planning under uncertainty in PEC domains.
\end{itemize}

\section{Background}

\subsection{Probablistic Event Calculus}\label{sec:PECEPECbackground}
This section provides a compact overview of the key components of PEC relevant to this work. The interested reader is referred to \cite{d2017foundations} for a more complete treatment. The Probabilistic Event Calculus (PEC) is an action-language framework for representing and reasoning about the effects of actions over time in domains with inherent uncertainty, combining a formal logical structure with probabilistic modelling. A PEC domain $\pecdomain$ comprises a finite, non-empty set of fluents $\mathcal{F}$ and values $\mathcal{V}$, a finite set of actions $\pecactions$, and a non-empty set of time instants $\mathcal{I}$. Fluents, as in classical narrative reasoning formalisms \cite{sandewall1992features, mccarthy1959programs}, refer to properties of the world which may be affected by actions taken. 

The function $\vals$  maps a fluent or action to the values that they can take; thus, $\pecfluent \in \pecfluents$ can take values in $\vals(F)$. Note that all actions $\pecaction\in \pecactions$ are restricted such that $\vals(A) = \{\top,\bot\}$ indicating whether or not they are performed. A literal $\literal\in\literals$ is either a fluent literal or an action literal. A fluent literal $\literal_F\in \literals_F$ is an expression of the form $F =V$ for some $F \in \pecfluents$ and $V \in vals(F)$, while an action literal $\literal_\pecaction\in \literals_\pecaction$ is either $\pecaction = \top$ or $\pecaction = \bot$. A state $\pecstate \in \pecstates$ is an assignment of values to all fluents and actions through literals, a fluent state $\pecfluentstate \in \pecfluentstates$ is an assignment of values to all fluents of the domain through fluent literals, while a partial fluent state $\partialfluentstate \in \partialfluentstates$ is a subset of value-assigned fluents, i.e., $\partialfluentstate\subseteq\pecfluentstate$ for some $\pecfluentstate\in\pecfluentstates$. 

PEC uses a set of action-language style propositions to specify probabilistic narrative information. A domain consists of: 
\begin{enumerate}
    \item A finite set of \textit{v-propositions} which detail the values that a fluent may take. The v-proposition takes the form: $$F \textbf{ takes-values } \{V_1, V_2, ..., V_m\}$$
    \item Exactly one \textit{i-proposition} which specifies the probabilities of initial fluent states that hold at the minimum time instant. 
    This takes the form:
    $$\textbf{ initially-one-of  } \{(\pecfluentstate_1,P_1^+),\ldots,(\pecfluentstate_J,P_J^+)\}$$
    where $\pecfluentstate_j$ is the fluent state associated with outcome $j$ and $P_j$ is its associated probability.
    \item A finite set of \textit{c-propositions} which model the causal effects of actions, taking the form:
    $$\theta \textbf{ causes-one-of } \{(\partialfluentstate_1,P_1),\ldots,(\partialfluentstate_J,P_J)\}$$
    where the \emph{body} $\theta$ is a formula closed under conjunction and negation which specifies the preconditions for causal effects to take hold, such that at least one action is true; while the \emph{head} describes the possible outcomes and where $\partialfluentstate_j$ is the partial fluent state associated with outcome $j$ and $P_j$ is its associated probability. An outcome updates the fluent state to the fluent value assignments in the partial state.
    \item A set of \textit{p-propositions} modelling the occurrence of actions, taking the form:
    $$A \textbf{ performed-at } I \textbf{ with-probs } P^+ \textbf{ if-holds } \tilde{X}$$
    which states that action $A$ takes value $\top$ at instant $I$ with probability $P^+$ if partial fluent state $\tilde{X}$ holds at $I$.
\end{enumerate}

A PEC domain description, $\pecdomain$, is then a collection of i-, v-, c-, and p-propositions. Note that, for any two distinct c-propositions in $\pecdomain$ with bodies $\theta_1$ and $\theta_2$, no state $S$ can entail both $\theta_1$ and $\theta_2$; this constraint ensures that at each instant, an environment can be affected by only one set of preconditions, eliminating the possibility of contrasting sets of effects. 

PEC supports a possible-worlds semantics, where a world is an evolution of an environment over time. Using this semantics, \emph{temporal projection} computes the probability of fluent states or partial fluent states at future time points given an initial state distribution and a narrative of action occurrences. The original PEC implementation uses answer set programming (ASP) to enumerate all ``well-behaved'' worlds with respect to a domain $\pecdomain$. The probability of a query about a partial fluent state $\partialfluentstate$ at some instant $I$ is calculated by summing the probabilities of all well-behaved worlds where $\partialfluentstate$ holds at instant $I$. This original implementation faces scalability challenges as domain complexity increases, leading to exponential growth in possible worlds. Consequently, alternative approaches emerged: an Anglican-based method for approximate estimation \cite{dasaro2019phd}, and an iterative approach \cite{d2023application} for exact computation.

\subsection{Markov Decision Processes}\label{sec:MDPs}
An MDP is defined as the tuple $(\mdpstates, \mdpactions, \mdpinitdist, \transfunc, R)$, where $\mdpstates$ and $\mdpactions$ are finite sets of \textit{states} and \textit{actions} respectively, where the sets $\mdpstates$ and $\mdpactions$ in this context are not to be confused with PEC states and actions. Typically, states and action are treated as unitary entities without internal structure. The initial state distribution is given by $p_0$, where the probability of starting at time $t=0$ in state $s$ is given by $\mdpinitdist(s)$. Transition dynamics are given by transition function $\transfunc$ which describes for any time $t$ the probability distribution over next state, $s'$, given current state, $s$, and action taken, $a$, i.e.
$$\transfunc(s,a,s') =\Pr(s_{t+1}=s'|a_{t}=a,s_{t} = s)$$

The reward function $R$ maps each state transitions to a numerical rewards, so if $r$ is the reward received from transitioning from $s$ to $s'$ by taking action $a$ then:
$$R(s,a,s') = r$$

Behaviour in an MDP are encoded in a policy $\mu$. Most commonly, this is a \emph{stationary policy}, meaning it maps from states to actions, or probability distributions over actions, independent of time. 
Policies may be deterministic: $\mu(s)=a$, meaning that an action, $a$, is selected given state, $s$, or stochastic: $\mu(s,a) = \Pr(a_{t}=a| s_{t}=s)$, meaning that action, $a$, is probabilistically selected given state, $s$. 
More generally still, non-stationary MDPs \cite{lecarpentier2019non} allow for a policy which can be different for different points in time, so for $t\neq t'$ then $\mu_t(s,a)$ can be different from $\mu_{t'}(s,a)$.

A relevant extension to MDPs, Semi-Markov Decision Processes (SMDPs) \cite{bradtke1994reinforcement} allowing state transitions to occur after arbitrary real valued time intervals, with actions triggering transitions according to probability distributions over sojourn times (the interval during which an action persists until a transition occurs). The options framework \cite{sutton1998between} can be viewed as the middle-ground between MDPs and SMDPs. This provides a hierarchical framework where temporally-extended actions (termed options) can be executed over variable but discrete number of time steps, with each option having its own policy for selecting primitive actions until some termination condition is met.

\section{Comparison of Related Frameworks}\label{sec:pec comparison}

This section compares the PEC framework to the MDP, and the two MDP variants described above which hold similar modelling properties to PEC. Generally, both PEC and MDPs model sequences by defining key narrative elements: i. a finite set of actions and variables to encode the changing environmental state, ii. a probabilistic distribution of initial environmental states, iii. sets of probabilistic state dependent, time independent, effects of actions on the environment, and iv. behavioural rules describing state dependent action occurrences (p-propositions for PEC and policies for MDPs).

\begin{table*}[t]
\centering
\renewcommand{\arraystretch}{1.45}
\caption{Comparison of Key Attributes between related frameworks.}
\scriptsize

\begin{tabular}{|p{1.7cm}|p{2.5cm}|p{2.5cm}|p{2.5cm}|p{2.5cm}|}
\hline
\textbf{Property} & \textbf{PEC} \cite{d2017foundations} & \textbf{MDP} & \textbf{Semi-MDP} \cite{bradtke1994reinforcement} & \textbf{Options} \cite{sutton1998between}\\
\hline
Time & Discrete & Discrete & Continuous & Discrete \\
\hline
State & Fluent assignment & State variable & State variable & State variable\\
\hline
Action \mbox{Duration} & Uniform and discrete & Uniform and discrete & Variable and continuous & Variable and discrete\\
\hline
Time Between Actions & Variable and discrete & Uniform and discrete & Variable and continuous & Variable and discrete\\
\hline
Simultaneous Actions & Yes & No & No & No\\
\hline
Action-Transition Relation & Action-taking independent to state transitions & Action-taking synchronised with state-transitions & Action-taking synchronised with state-transitions & Action-taking synchronised with state-transitions\\
\hline
Decision-making & At specified time instants with fluent conditions & Policy mapping states to actions & Policy mapping states to actions & Policy mapping states to options\\
\hline
\end{tabular}
\label{tab:pec_comparison}
\end{table*}

One of the key distinctions between PEC and MDP formalisms lies in their representation of the environment. PEC represents systems through value-taking fluents as properties, offering a structured representation. In contrast, MDPs explicitly represent each possible situation as an atomic state value. Consequently, MDPs operate on transitions between complete states, while PEC reasons about changes in specific system properties. 

Another fundamental difference is the existence of a reward mechanism in MDPs, which is absent in PEC. This reflects PEC's primary focus on temporal reasoning without any encoding of preferences over action-taking and outcomes. MDPs, in contrast, originated in the study of stochastic optimal control and have become a natural foundation for reinforcement learning problems where behaviour is reinforced with respect to some reward criteria.

The two frameworks both adopt a single temporal scale for action occurrence such that the effects of an action at $t$ are assumed to always take hold at $t+1$. However, the abstraction of time in relation to action-taking is fundamentally different in the two frameworks. Where PEC represents the progression of time independently to actions and environmental changes, the MDP assumes a direct correspondence of each discrete time step to each episode of agent-state interaction. As a result, the time between action-taking points may be variable in PEC but are assumed uniform in the MDP. 

Action-taking in PEC conditions obligatorily on time-instants and optionally on specific partial fluent conditions. This differs from the MDP's standard stationary policy which conditions only on the state an agent resides in. A non-stationary policy \cite{lecarpentier2019non} does incorporate a temporal component with a time-evolving probability distribution over actions in each state, aligning more closely to PEC. Nevertheless, the non-stationary policy is primarily used for non-stationary MDPs with time-evolving transition and reward mechanisms, where the optimal policy may not be stationary. 

The SMPD \cite{bradtke1994reinforcement} models temporally extended actions through probabilistic distributions over sojourn times. Diverging from PEC and the MDP, the SMPD operates in a continuous-time system which allows for variable and continuous-time scales for action-occurrence. As a result, it is similar to PEC in allowing variable intervals between decision points. However, action-taking in the SMDP is still strictly tied to state-transitions when the system enters a new state. Thus, the interval variability between decision points in the SMPD arises from its probabilistic action duration mechanisms; in contrast, the variability of decision points in PEC is a result of its highly flexible system of action-taking, independent to time-flow.
 
The options framework \cite{sutton1998between}, building on Bradke \& Duff's SMPD methods \cite{bradtke1994reinforcement}, generalises actions to temporally extended courses of action within an underlying discrete-time MDP. It is similar to PEC in its ability to model variable, discrete intervals between decision points. However, action-taking is still strictly bound to transition points such that options on each hierarchy are conducted contiguously. This framework enables options to be hierarchically nested, in contrast to PEC and MDPs which maintain a flat representation of action-taking. 

Generally, the independent flow of time in PEC reflects its assumptions of default persistence-- where properties maintain their values and actions remain inactive unless explicitly specified or triggered. This is in contrast to MDP frameworks model which model stochastic processes that are constantly undergoing change. As a result, PEC's model of action-taking in relation to time is highly flexible, allowing for both simultaneous actions or no actions at a given time instant. While MDP variants allow for variable time between actions, which might appear similar to PEC's temporal flexibility on the surface, they still adhere to a strict correspondence between state transitions and action execution that PEC does not require. The aforementioned differences are summarised in Table \ref{tab:pec_comparison}.

\section{The PEC-MDP Formalism and its Applications}

\subsection{MDP dynamics}

This section presents a formal translation of PEC to the PEC-MDP, an MDP derivative without rewards. As discussed in section~\ref{sec:pec comparison}, PEC shared similarities with the options framework in that both allow for discrete and variable intervals between actions. However, rather than adopting the options framework, which primarily serves hierarchical learning purposes outside PEC's scope, we will modify the standard MDP to accommodate PEC's more flexible action-taking mechanisms. To this end, we introduce action-taking situations which includes \emph{single}, \emph{composite}, and \emph{null} actions to simulate time-points at which agents cannot act or performs multiple actions simultaneously. 
Furthermore, given that PEC's action-taking is conditioned on time, non-stationary policies will be adopted as they allow for time-dependent policies while maintaining stationary transition dynamics. The reward component of the MDP framework is omitted in the initial translation, as PEC domains are without inherent reward signals. The details follow.

\textbf{Time steps.}\quad We define the function $\timemap$ to normalise the PEC timeline to start at 0 while preserving temporal ordering. Thus, $\timemap:\instants \to \mdptimes$, where $\delta(I_1) \leq \delta(I_2)$ iff $I_1 \leq I_2$, and $\mdptimes = \{0,\ldots,\numinstants-1\}$ is the set of time steps in the PEC-MDP, where $\numinstants = |\instants|$.

\textbf{States.}\quad We propose a two-step translation process of PEC fluent states to the PEC-MDP state representation. First, the PEC fluent state $\tilde{S}$ is mapped to a vector representation $x$, obtaining a purely numerical encoding of fluent values. Each unique PEC-MDP vector representation is then mapped to an integer representation $s$, for a more compact representation supporting table-lookup operations (common in MDP frameworks).

For the vector state representation, we assume a canonical ordering of PEC fluents,
$\pecfluents = \{\orderedelement{F}{0}, \ldots , \orderedelement{F}{\numfluents-1}\}$, where $\numfluents = \cardinalityof{\pecfluents}$,
a similar ordering over PEC values,
$\pecvalues = \{\orderedelement{V}{0},\ldots, \orderedelement{V}{\numvalues-1}\}$, where $\numvalues = \cardinalityof{\pecvalues}$. We propose an indexing function for values, $\valueindexof:\pecvalues \to \{0,\ldots,\numvalues-1\}$ where $\valueindexof(\orderedelement{V}{i}) = i$. A fluent state, $\pecfluentstate$, in PEC can then be mapped to vector of value indices, as
$$
x(\pecfluentstate)= (\valueindexof(V_0), \ldots ,\valueindexof(V_{\numfluents-1})) $$
where $\forall i \,:\, (F^{(i)}=V_i) \in \tilde{S}$. Let $\vectorrepset$ denote the complete set of all possible vector state representations so $\vectorrepset = \{x(\tilde{S})\,:\, \tilde{S} \in \mathcal{\tilde{S}} \}$ with $\numvecreps = \cardinalityof{\vectorrepset}$.

Our second step then maps each vector representation $x$ to an integer via indexing function $\vectorrepindexof:\vectorrepset\to \{0,\ldots,\numstates\}$. To achieve this, we first define a preference ordering, $\prec$, over states given by:
$$x \prec x' \,\,\, \Leftrightarrow \exists \, K \,:\, \forall \, k \, < \, K \,:\, x_k = x_k' \text{ and } x_{K} < x_{K}'$$
Here, $x_k$ is the $k$th element of $x$ and $K \in \{0,\ldots,\numfluents-1\}$.
We then inductively define $\vectorrepindexof$ as:

$$
\vectorrepindexof(x) = 
\begin{cases}
0 & \text{if } \forall x' \in \vectorrepset \text{ with } x \neq x': x \prec x' \\
\vectorrepindexof(x')+1 & \text{if } \exists x' \in \vectorrepset \text{ s.t. } 
    x' \prec x \; \text{and} \\
    & \not\exists x'' \in \vectorrepset\,:\, x' \prec x'' \wedge x'' \prec x
\end{cases}
$$

Elements of $\vectorrepset$ are thus mapped to their corresponding position in the ordering. For brevity, we denote the resulting set of integer states $\mdpstates = \{\vectorrepindexof(x) \,:\, x \in \vectorrepset\}$, and define fluent state indexing function $\stateindexof:\pecfluentstates\to \mdpstates$ for all $\pecfluentstate\in \pecfluentstates$ as $\stateindexof(\pecfluentstate) = \vectorrepindexof(x(\pecfluentstate))$. 

\textbf{Actions.} 
Action-taking situations may refer to a single action, a composite of concurrent actions, or the absence of an action. To delineate the total set of possible action-taking situations in a domain, we define a set of actions that may be performed at any time instant $I$: $\possibleactionsat{I} = \{\pecaction  \,:\, (\pecaction \textbf{ performed-at } I)\in \pecdomain\}$.
All action-taking situations at a given instant $I$ may be denoted as the powerset of $\possibleactionsat{I}$, denoted $\powersetof{\possibleactionsat{I}}$.
The overall set of action-taking situations is the set containing all such sets: $\actionsituations = \{\mathcal{P}(\pecactions_I) \,:\, I\in \mathcal{I}\}$. $\actionsituations$ thus contains all possible single or composite (simultaneous) actions that may be taken in domain $\pecdomain$ as well as the null (do-nothing) action $\nullactionsituation$.
We define an arbitrary ordering over $\actionsituations$ denoted $\mathcal{A}_u = \{\orderedelement{\actionsituation}{0}, \orderedelement{\actionsituation}{1}, ..., \orderedelement{\actionsituation}{M-1} \}$, and the indexing function $\situationindexof:\actionsituations \to \{0,\ldots,\numsituations-1\}$ where $\situationindexof(\orderedelement{A_u}{i}) = i$ and $\numsituations=\cardinalityof{\actionsituations}$. Let $\encodedsetsituations$ denote the set containing all integer encoded action-taking situations of a domain, with a single element of this set denoted as $a$. For convenience, the element referring to the null-action situation is denoted $\nullaction = \situationindexof(\nullactionsituation)$.

\textbf{Initial Distribution.} We map the PEC i-proposition outcomes $\{(\pecfluentstate_1,P_1^+),\ldots,(\pecfluentstate_J,P_J^+)\}$ to the MDP probability function $p_0$. For fluent states not outlined in a potential outcome, we set its corresponding probability to $0$.
$$p(s_0 = \stateindexof(\pecfluentstate)) = 
\begin{cases}
   P_i^+ & \text{if } \exists (\pecfluentstate_i,P_i^+) : \pecfluentstate = \pecfluentstate_i \\
   0 & \text{otherwise}
\end{cases}$$

\textbf{Transition Function.} We will now translate the causal effects delineated by c-propositions to the MDP transition function, $\transfunc$. It is first necessary to extract any fluent preconditions and the action(s) from the body of a c-proposition $\theta$. To do this, we establish two components: the action component $\bodyactions = \{\pecaction \in \pecactions \,:\, \pecaction = \top \text{ is a constitute of } \theta\}$ and the fluent preconditions $\tilde{X}_\theta = \{\literal \in \literals_F \,:\, \literal \text{ is a constitute of } \theta\} \in \tilde{\mathcal{X}}$. $\bodyactions$ is the action-taking situation indicated by $\theta$ where $\bodyactions \in \actionsituations$ and a fluent state meets the preconditions of $\theta$ if and only if it entails $\partialfluentstate_\theta$. Thus, the set of all fluent states that satisfy body $\theta$ are denoted as $\bodystates = \{\pecfluentstate \,:\,\pecfluentstate \models \tilde{X}_\theta\}$. It can be noted that where there are no fluent preconditions in $\theta$, any fluent state satisfies this empty set of conditions. We denote the PEC-MDP action corresponding to this action-taking situation as $a_\theta = \situationindexof(\bodyactions)$, and the associated set of PEC-MDP states as $\mdpstates_\theta = \{s\,:\, \pecfluentstate \in \bodystates, s=\stateindexof(\pecfluentstate)\}$.

We can now examine the c-proposition head $\{(\partialfluentstate_1,P_1),\ldots,(\partialfluentstate_J,P_J)\}$ to obtain state transitions. The effects of an outcome detailed by $\partialfluentstate_i$ must be integrated with the fluent state that corresponds to the current PEC-MDP state. 
To implement this in our vector representation, we define a function $\Upsilon$ that takes a partial fluent state $\partialfluentstate$ and a vector representation $x = (v_0, v_1, \ldots, v_{\numfluents-1})$ of a fluent state, and produces an updated vector representation:
$$\Upsilon(\partialfluentstate, x) = (v'_0, v'_1, \ldots, v'_{\numfluents-1})$$
\noindent where each element $v'_i$ is determined as follows:
$$v'_i = 
\begin{cases}
\valueindexof(V_i)   & \text{if } (F^{(i)}=V_i) \in \partialfluentstate \\
v_i    & \text{otherwise.}
\end{cases}$$
This operation updates only those fluent values specified in the partial fluent state $\partialfluentstate$, while preserving all other values from the original state vector.
We can now define probabilities in our PEC-MDP transition function $T(s,a,s')$. For each tuple $(s,a)$ we must determine whether the state $s$ and action $a$ satisfies a body of any c-proposition in the original PEC. If there is no such c-proposition, including when $a = \nullaction$, then we define:
$$
T(s,a,s') = 
\begin{cases}
1 & \text{if } s'=s \\
0 & \text{otherwise.}
\end{cases}
$$
If there is such a body $\theta$, where $s \in \mdpstates_\theta$ and $a = a_\theta$. We define the outcomes of that c-proposition as $\{(\partialfluentstate_1,P^+_1),\ldots,(\partialfluentstate_J,P^+_J)\}$ and associate current and next state with their state representations $s = \stateindexof(x)$ and $s' = \stateindexof(x')$ and say:
$$
T(s,a,s') = T(\stateindexof(x),a,\stateindexof(x'))= \sum_{\partialfluentstate_j \text{s.t.} \Upsilon(\partialfluentstate_j, x)=x' } P^+_j
$$
In words, the probability of transitioning to $s'$ from $s$ under action $a$, is the sum of all outcomes that update $s$ to be $s'$.

\textbf{Policy.} This translation adapts a non-stationary MDP policy to capture the conditional probabilities of actions taken in PEC. We define the policy as $\mu(a,s,t) = p(a|s,t)$, where $t$ is an independent input, reflecting PEC's time-conditioned actions. Due to PEC's flexible action-taking, this policy represents a distribution over action-taking situations $\actionsituation\in \actionsituations$ rather than individual actions.

First, we define an arbitrary ordering over actions in $\pecactions$ such that $\pecactions = \{\orderedelement{A}{0}, \orderedelement{A}{1}, ..., \orderedelement{A}{\numactions-1}\}$ where $\numactions = \cardinalityof{\pecactions}$. To extract probabilities from p-propositions, we define a probability matrix $\orderedelement{P}{I} \in \mathbb{R}^{\numstates \times \numactions}$ for each time instant $I\in \mathcal{I}$, where each entry $p_{j,k}^{(I)}$ represents the probability of action $\orderedelement{A}{k}$ occurring at time $I$ in state $\orderedelement{\pecfluentstate}{j}$:
$$
p_{j,k}^{(I)}= \begin{cases} 
    P^+ & \text{if } \exists \text{ p-proposition } (\orderedelement{A}{k} \textbf{ performed-at } I 
         \textbf{ with-prob } P^+ 
         \textbf{ if-holds } \partialfluentstate)\in \pecdomain \\
        & \text{such that } \orderedelement{\pecfluentstate}{j} \models \partialfluentstate \\
    0   & \text{otherwise}
\end{cases}
$$
For each action-taking situation $\actionsituation \in \actionsituations$, we calculate the probability of its occurrence at instant $I$ in state $\orderedelement{\pecfluentstate}{j}$ as follows:
\begin{align}
P(\actionsituation | I, \orderedelement{\pecfluentstate}{j}) = 
\begin{cases}
\displaystyle\prod_{\orderedelement{A}{k} \in \actionsituation} p_{j,k}^{(I)} \cdot \displaystyle\prod_{\orderedelement{A}{i} \in \pecactions \setminus \actionsituation} (1-p_{j,i}^{(I)}) & \text{if } \actionsituation \neq \emptyset \\[6ex]
\displaystyle\prod_{k=0}^{\numactions-1} (1-p_{j,k}^{(I)}) & \text{if } \actionsituation = \emptyset
\end{cases}
\notag
\end{align}
In the first case (where the action-taking situation is not empty), the probability is calculated as the product of:
\begin{itemize}
    \item The probability that each action in situation $\actionsituation$ will be taken
    \item The probability that each action not in situation $\actionsituation$ will not be taken
\end{itemize}
For the null-action situation ($\actionsituation = \emptyset$), the probability is the product of the probabilities that each possible action will not be taken.

Finally, we define the policy function $\mu$ for our PEC-MDP as:
$$\mu(a,s,t) = P(\situationindexof^{-1}(a) \,| \, \timemap^{-1}(t), \stateindexof^{-1}(s))$$
This maps our encoded MDP state $s$, action $a$, and time $t$ to the probability of the corresponding action-taking situation occurring at the corresponding PEC instant and fluent state.\\

In summary, the PEC-MDP uses numerical 0-based encoding of all valued PEC components including time instants, actions, fluents, and values, which replaces PEC's natural language representations for better efficiency and scalability, particularly in complex domains. Transition probabilities and policies can be easily stored in matrices with indices referring to states, actions, or times. The main trade-off of this approach is the reduced interpretability of PEC variables. However, this is addressed by storing mappings that allow elements to be translated between each form of representation. State, value, time instants, and initial distribution encoding functions are entirely bijective, while transition and action-taking translation functions are not so, given that we aggregate over transition probabilities and action probabilities for action-taking situations. As we are primarily interested in fluent values through time or action-taking mechanisms, it is less of a concern for transition mechanisms to be fully reconstructable in the original domain. However, the ability to translate action-taking mechanisms back to PEC's readable format is crucial for interpretable optimisation, which will be discussed in Section~\ref{sec:RL} when we explore objective-directed strategies.



The PEC-MDP translation has been fully implemented in Python and may be found on the \href{https://github.com/LyrisX02/PEC-MDP}{PEC-MDP} GitHub repository. The code is structured following the object-oriented paradigm within the main module \href{https://github.com/LyrisX02/PEC-MDP/blob/main/PEC_Parser.py}{PEC-parser}. For a comprehensive overview of the repository structure, example domains, and usage instructions, readers are directed to the repository's \href{https://github.com/LyrisX02/PEC-MDP/blob/main/README.md}{README file}. 



\subsection{Temporal Projection}\label{sec:temp}

Temporal projection is a fundamental reasoning task for narrative domains, calculating the probability that certain properties of a system will hold at specified time points given an initial state distribution and an action narrative. Specifically in PEC, this process traces how fluent values evolve over time in response to action occurrences, while accounting for the probabilistic effects encoded in the domain description. The basic form may be formulated as the query: ``Based on the given initial distribution over states, what is the probability of $\tilde{X}$ holding at instant $I$?''. 


To compute the probability that some partial fluent state $\partialfluentstate_{\query}$ holds at time $I_{\query}$, we first identify the set of fluent states that satisfy the partial fluent state as $\pecfluentstates_{\query} = \{\pecfluentstate \in \pecfluentstates \mid \pecfluentstate \models \partialfluentstate_{\query}\}$ and the PEC-MDP time equivalent of the queried instant as $t_{\query}=\timemap(I_{\query})$. Define $\mathbf{p}_t \in [0,1]^{n_s}$ as the distribution over states at time $t$, and $M_t$ is the policy-weighted transition matrix at time $t$ defined as\footnote{We use $[\cdot]_{i}$ and $[\cdot]_{i,j}$ to indicate the $i$th element of a vector and $(i,j)$-element of a matrix respectively. }:
$$[\mathbf{M}_t]_{s,s'} = \sum_{a \in \mdpactions} \mu(a,s,t) \cdot T(s,a,s')$$
This matrix effectively combines the action selection probabilities with transition dynamics, enabling the direct propagation of state probabilities through time. The state distribution at time $t$ can be calculated iteratively based on the distribution at time $t-1$ as $\mathbf{p}_t = \mathbf{p}_{t-1}^T \mathbf{M}_{t-1}$. Thus, starting from the initial distribution $\mathbf{p}_0$, the state probability distribution may be propagated forward through successive applications of the policy-weighted transition matrix until time $t_{\query}$ to retrieve the state distribution at the queried time as, 
\begin{equation}\mathbf{p}_{t_{\query}} = \mathbf{p}_{0}^T\prod_{\tau=0}^{t_{\query}-1} \mathbf{M}_{\tau}
\qquad (\text{for stationary policies } \mathbf{M}_t = \mathbf{M} \text{ for all }t)
\label{eqn:distribution_at_t}
\end{equation}
The probability that $\partialfluentstate_{\query}$ holds at $t_{\query}$ is then computed by summing over the probabilities of all fluent states that entail $\partialfluentstate_{\query}$, i.e. $$P(\partialfluentstate_{\query} \,@ \,t_{\query}) = \|\mathbf{p}_{t_{\query}} \odot \mathbf{f}_{\query}\|_1$$  
where $\odot$ represents element-wise multiplication, $\|\cdot\|_1$ is the L1 norm, $\mathbf{f}_{\query}$ is a binary vector which is $1$ for all indices corresponding to fluent states entailed by $\partialfluentstate_{\query}$, i.e. $[\mathbf{f}_{\query}]_{s} = \indicatorfunc{\stateindexof(\pecfluentstate)=s \text{ for some }\pecfluentstate \models \partialfluentstate_{\query}}$, and $\indicatorfunc{\cdot}$ is the indicator function.

A more general form of temporal projection conditions this query on a prior partial fluent state and takes the form ``What is the probability of $\partialfluentstate_{\query}$ at instant $I_{\query}$, given $\partialfluentstate_{\cond}$ at earlier time $I_{\cond}$?''. 
To calculate this, we must first identify $\pecfluentstates_{\query}$ and $I_{\query}$ as before, as well as the set of fluent states satisfying the condition, i.e. $\pecfluentstate_{\cond} = \{\pecfluentstate \in \pecfluentstates \mid \pecfluentstate \models \partialfluentstate_{\cond}\}$, and the associated time $t_{\cond}=\timemap(I_{\cond})$. We calculate the unconditioned state distribution at $t_\cond$, $\mathbf{p}_{t_\cond}$, using Equation \eqref{eqn:distribution_at_t}.
Let $\mathbf{f}_{\cond}$ be a binary vector playing the same role for $\partialfluentstate_{\cond}$, as $\mathbf{f}_{\query}$ plays for $\partialfluentstate_{\query}$.
We can then compute distribution at time $t_{\cond}$ given that partial fluent state $\partialfluentstate_{\cond}$ holds as:
$$\mathbf{p}_{t_{\cond}|\partialfluentstate_{\cond}} = \frac{\mathbf{p}_{t_{\cond}} \odot \mathbf{f}_{\cond}}{\|\mathbf{p}_{t_{\cond}} \odot \mathbf{f}_{\cond}\|_1}$$
This ensures the distribution is properly normalised. 
We can then propagate this conditional distribution forward to time $t_{\query} > t_{\cond}$ through $t_{\query} - t_{\cond}$ iterative applications of the transition matrices and apply filter vector $\mathbf{f}_{\query}$ to give: 
$$P(\tilde{X}_{\query}@t_{\query} \mid \tilde{X}_{\cond}@t_{\cond})
= \left\|\left(\mathbf{p}_{t_\cond|\partialfluentstate_\cond}^T\prod_{\tau=t_{\cond}}^{t_{\query}-1}\mathbf{M}_{\tau} \right) \odot \mathbf{f}_{\query}\right\|_1
$$
This vector-based approach provides efficient computation for both basic and conditional temporal projection queries without requiring explicit enumeration of all possible worlds. We implemented these temporal projection capabilities in the PEC-MDP repository within the \href{https://github.com/LyrisX02/PEC-MDP/blob/main/temporal_projection.py}{temporal projection module}. This module can be used in conjunction with the PEC-Parser to directly perform temporal projection queries on raw PEC domain strings.

\subsection{Objective-directed Strategies for PEC}\label{sec:RL}

Narrative domains often involve agents making purposeful decisions to achieve specific goals rather than merely following predefined action sequences. While traditional PEC formulations excel at describing probabilistic narratives and their possible evolutions, they lack mechanisms for identifying optimal action policies when agents pursue objectives. The PEC-MDP formalism presents an opportunity to leverage the large body of research in reinforcement learning to compute objective-directed strategies within PEC domains. 

Naturally, a desirability criterion must first be established to guide agent behaviour. This criterion defines what constitutes success in the domain and may be translated into the MDP reward function. For example, in a medical treatment domain, a desirability criterion might prioritise patient recovery while minimising side effects; in a logistics scenario, it might balance delivery speed against operational costs.

Formally, the standard MDP reward function maps state-action-state transitions to scalar reward values. This general formulation allows for capturing a variety of desirability criteria when applied to PEC domains. In PEC narrative contexts, common desirability criteria may include:
\begin{itemize}
   \item \textbf{Goal achievement}: When the objective involves reaching states satisfying certain fluent conditions $\partialfluentstate_{\text{goal}}$, we identify the set of goal states $\pecfluentstates_{\text{goal}} = \{\pecfluentstate \in \pecfluentstates \mid \pecfluentstate \models \partialfluentstate_{\text{goal}}\}$. We then define $R(s, a, s') = r_{\text{goal}}$ when $s' \in \{\stateindexof(\pecfluentstate) \mid \pecfluentstate \in \pecfluentstates_{\text{goal}}\}$. Otherwise set $R(s,a,s')=0$.

   \textbf{Action costs}: When narrative domains involve resource expenditure or risk associated with actions, negative rewards may be attached to individual PEC actions $\pecaction \in \pecactions$. We first define costs at the PEC action level: $c(\pecaction)$ represents the cost associated with performing action $\pecaction$. These action-level costs are then mapped to MDP action-taking situations to create the reward function: $R(s, a, s') = R(s') - C(a)$ where $C(a)$ aggregates the costs of all actions in the situation $\actionsituation = \situationindexof^{-1}(a)$. For example, a natural aggregation would be: $C(a) = \sum_{\pecaction \in \situationindexof^{-1}(a)} c(\pecaction)$. This approach ensures that costs defined at the intuitive PEC action level are properly translated to the MDP's action-taking situations.


   \item \textbf{Temporal efficiency}: When timely achievement of goals matters, a constant negative reward per step can be applied, encouraging efficient narratives without changing the fundamental optimal policy. This may also be reinforced through the discount factor in standard RL algorithms \cite{amit2020discount}.

\end{itemize}
We will not delve further into exact reward function mappings as these are highly domain-specific. Once an appropriate quantitative reward signal is defined over outcomes, actions, or transitions, suitable reinforcement learning methods can be applied to discover optimal policies for the narrative domain.


To preserve the interpretability advantages of PEC's original formalism, we present a method for translating learned policies back into human-readable p-propositions. This translation works for both stationary and non-stationary policies, but requires deterministic policies (which always exists as an optimal solution for MDPs due to the Markov property). 
This is because a probabilistic policy in the PEC-MDP may assign probabilities to different action-taking situations at a single time point; given that PEC's p-propositions can only specify independent probabilities for individual actions, PEC cannot preserve the probabilistic relationships between actions in the discovered policy. 

To translate a deterministic stationary policy, $\mu: \mdpstates \rightarrow \mdpactions$, 
we construct corresponding p-propositions as follows: for each state $s \in \mdpstates$ where $\mu(s) = a$ and $a \neq a_\emptyset$ (not the null action), we identify the fluent state $\tilde{S}$ corresponding to $s$, as $\tilde{S} = \stateindexof^{-1}(s)$ and the set of individual PEC actions contained in the action-taking situation $a$, as $\{A_1, A_2, ..., A_k\} = \situationindexof^{-1}(a)$. For each PEC action $A_i$ in this set, we create a p-proposition: $$A_i \textbf{ performed-at } I \textbf{ with-prob } 1 \textbf{ if-holds } \tilde{S}$$
This is done for every time instant $I \in \mathcal{I}$ in the PEC domain, as the policy is stationary and thus applies the same state-action mapping regardless of time. 

Since this translation generates a large number of p-propositions (potentially $|\mathcal{I}| \times |\mdpstates| \times k$ where $k$ is the average number of atomic actions per action-taking situation), refinements can be applied to reduce this set for interpretability while maintaining semantic equivalence. These include eliminating p-propositions for unreachable state-time combinations and generalising fluent state conditions to their minimal distinguishing features.

First, to identify which states are actually reachable at each time point, we perform a policy-guided Markov chain reachability analysis by computing the probability distribution over states at each time step. This may be done by using equation \eqref{eqn:distribution_at_t} up to the time step where the last action is performed. We can then eliminate p-propositions associated with states that have zero (or negligibly small) probability of being reached at specific time steps, significantly reducing the number of p-propositions in the final representation.

Second, fluent conditions in p-propositions need not specify complete fluent states. 
We can abstract each complete fluent state to a minimally strict set of fluent preconditions by identifying the smallest subset of fluents necessary to distinguish between all actions at each time step. For each time step, let $S_t = \{s_1, s_2, ..., s_n\}$ be the set of all states appearing in p-propositions at that time step. For each state $s_i$ with corresponding vector representation $x_i$, we seek a minimal partial state representation $\hat{x}_i \subseteq x_i$ that distinguishes it from all other states in $S_t$ while using the fewest fluent assignments possible. Formally, $\hat{x}_i$ must satisfy two conditions: (1) no other state $s_j \in S_t$ with $j \neq i$ is compatible with all fluent assignments in $\hat{x}_i$, and (2) removing any single fluent assignment from $\hat{x}_i$ would violate condition (1).

Through these refinements, the translation produces a set of p-propositions that is not only functionally equivalent to the learned stationary policy but also refined for interpretability within the PEC framework.

For non-stationary policies, $\mu: \mdpstates \times \mdptimes \rightarrow \mdpactions$, the translation follows a similar approach but respects the time-dependence. Instead of applying the same mapping for every time instant, we create p-propositions specific to each time point: for each tuple $(s, t)$ where $\mu(s, t) = a$ and $a \neq a_\emptyset$, we identify $\tilde{S} = \stateindexof^{-1}(s)$, the PEC time instant $I = \timemap^{-1}(t)$, and the set of actions $\{A_1, A_2, ..., A_k\} = \situationindexof^{-1}(a)$. For each $A_i$, we create the p-proposition: $$A_i \textbf{ performed-at } I \textbf{ with-prob } 1 \textbf{ if-holds } \tilde{S}$$
The same refinement techniques for state reachability analysis and minimal fluent condition identification can be applied to non-stationary policies as well, with the additional benefit that the time-specific nature of the policy often naturally reduces the number of required p-propositions.

The PEC-MDP repository includes a Python implementation that enables users to convert learned policies back to p-propositions through the \href{https://github.com/LyrisX02/PEC-MDP/blob/main/policy_to_pprops.py}{policy translation module}. Additionally, it demonstrates the successful application of reinforcement learning on a PEC logistics domain in the \href{https://github.com/LyrisX02/PEC-MDP/blob/main/boxworld.ipynb}{BoxWorld demonstration notebook}, where an optimal policy was discovered and translated back to human-readable PEC notation.

\section{Conclusion}

This paper presented a comprehensive translation of the Probabilistic Event Calculus (PEC) into a Markov Decision Process (MDP) framework, termed the PEC-MDP. The translation process involved several key aspects:
\begin{enumerate}
    \item Developing a bidirectional numerical encoding scheme for PEC's domain elements, including fluents, values, actions, and time instants.
    \item Adapting PEC's probablistic components into probability functions for an initial state distribution, state transitions, and non-stationary action policies.
    \item Introducing the concept of ``action-taking situations'' to handle PEC's flexible action occurrence mechanism within the more rigid MDP framework.
\end{enumerate}
We explored how this translation enables both efficient temporal projection (Section~\ref{sec:temp}) and objective-directed reasoning through reinforcement learning (Section~\ref{sec:RL}).
 

Finally, let us note that other probabilistic extensions of the Event Calculus have been proposed, focusing on event recognition using probabilistic logic programming and learning from noisy data in both offline and online settings, i.e., \cite{skarlatidis2015mlnec, skarlatidis2015probabilistic, mantenoglou2023online, katzouris2023online,mantenoglou2024reasoning}. In contrast, our PEC-MDP framework inherits the main features of PEC, which is more expressive (see \cite{d2020probabilistic} for a comparison), compiling it into an MDP. This shift lays the groundwork for reinforcement learning, positioning our work toward goal-driven reasoning and policy optimisation rather than event detection.

While the current paper focuses on the PEC-MDP's application to temporal projection and objective-directed learning, the PEC-MDP framework lays the groundwork for further applications that leverage MDP-based techniques within narrative domains. Beyond these broader applications, future work will include formal efficiency analyses and extending the framework to the Epistemic Probabilistic Event Calculus (EPEC) \cite{d2020probabilistic}.

\bibliography{refs}

\end{document}